\documentclass{article}

\usepackage{microtype}
\usepackage{graphicx}
\usepackage{booktabs} 
\usepackage{caption}
\usepackage{subcaption}
\usepackage{cite}

\usepackage{hyperref}

\usepackage[accepted]{icml2018}
\usepackage{qiangstyle}

\newcommand{\pia}{{\pi}}
\newcommand{\pie}{{\pi_e}}

\icmltitlerunning{Learning to Explore with Meta-Policy Gradient}

\begin{document}

\twocolumn[
\icmltitle{Learning to Explore with Meta-Policy Gradient}


\begin{icmlauthorlist}
\icmlauthor{Tianbing Xu}{baidu}
\icmlauthor{Qiang Liu}{utexas}
\icmlauthor{Liang Zhao}{baidu}
\icmlauthor{Jian Peng}{uiuc}
\end{icmlauthorlist}

\icmlaffiliation{baidu}{Baidu Research, Sunnyvale, CA, USA}
\icmlaffiliation{utexas}{University of Texas at Austin, USA}
\icmlaffiliation{uiuc}{University of Illinois at Urbana-Champaign, USA}


\icmlkeywords{Machine Learning, Reinforcement Learning}

\vskip 0.3in
]

\printAffiliationsAndNotice{\icmlEqualContribution} 


\begin{abstract}
The performance of off-policy learning, including deep Q-learning and deep deterministic policy gradient (DDPG), critically depends on the choice of the exploration policy.
Existing exploration methods are mostly based on adding noise to the on-going actor policy and can only explore \emph{local} regions close to what the actor policy dictates.   
In this work, we develop a simple meta-policy gradient algorithm that allows us to adaptively learn the exploration policy in DDPG. 
Our algorithm allows us to train flexible exploration behaviors that are independent of the actor policy, yielding a \emph{global exploration} that significantly speeds up the learning process. 
With an extensive study, we show that our method significantly improves the sample-efficiency of DDPG on a variety of reinforcement learning tasks. 
\end{abstract}

\section{Introduction}

Recent advances in deep reinforcement learning (RL) have demonstrated significant applicability and strong performance in games \cite{mnih:dqn, silver:2017}, continuous control \cite{lillicrap:ddpg}, and robotics \cite{levine:2016}. Among them, deep neural networks, such as convolutional neural networks, are widely used as powerful functional approximators for extracting useful features and enabling complex decision making. For instance, in continuous control tasks, a policy that selects actions under certain state observation can be parameterized by a deep neural network that takes the current state observation as input and gives an action or a distribution of action as output. In
order to optimize such policies, various policy gradient methods \cite{mnih:2016, schulman:trpo, schulman:ppo, heess:2017}, including both off-policy and on-policy approaches, have been proposed. In particular, deterministic policy gradient method (DPG), which extends the discrete Q-learning algorithm for the continuous action spaces, exploits previous experience or off-policy data from a replay buffer and often achieves more desirable sample efficiency compared to most existing on-policy policy gradient algorithms. In the recent NIPS 2017 learning to run challenge, the deep deterministic policy gradient algorithm (DDPG) \cite{lillicrap:ddpg}, a variant of DPG, has been applied by almost all top-ranked teams and achieved a very compelling success in a high-dimensional continuous control problem, while on-policy algorithms, including TRPO \cite{schulman:trpo} and PPO \cite{schulman:ppo}, performed much worse with the same amount of data collected.

In contrast to deep Q-learning (DQN) \cite{mnih:dqn} which only learns a value function on a set of discrete actions, DDPG also parameterizes a deterministic policy to select a continuous action, thus avoiding the optimization in or the discretization of the continuous action space. As an off-policy actor-critic method, DDPG utilizes Bellman equation updates for the value function and the policy gradient descent to directly optimize the actor policy. Unlike DQN which often applies epsilon-greedy exploration on a set of discrete actions, more sophisticated continuous exploration in the high-dimensional continuous action space is required for DDPG. A common practice of exploration in DDPG is to add a uncorrelated Gaussian or a correlated Ornstein-Uhlenbeck (OU) process  \cite{Uhlenbeck:ou} to the action selected by the deterministic policy. The data collected by this exploration method is then added to a replay buffer used for DDPG training. However, in practice, Gaussian noises may be sub-optimal or misspecified, and hyper-parameters in the noise process are hard to tune. 

In this work, we introduce a meta-learning algorithm to directly learn an exploration policy to collect better experience data for DDPG training. Instead of using additive noises on actions, we parameterize a stochastic policy to generate data to construct the replay buffer for training the deterministic policy in the DDPG algorithm. This stochastic policy can be seen as an exploration policy or a teacher policy that gathers high-quality trajectories that enable better training of the current deterministic policy and the value function. To learn the exploration policy, we develop an on-policy policy gradient algorithm based on the training improvement of the deterministic policy. First, we obtain a collection of exploration data from the stochastic policy and then apply DDPG on this data-set to make updates of the value function and the deterministic policy. We then evaluate the updated deterministic policy and compute the improvement of these updates based on the data just collected by comparing to the previous policy. Therefore, the policy gradient of the stochastic policy can be computed using the deterministic policy improvement as the reward signal. This algorithm adaptively adjusts the exploration policy to generate effective training data for training the deterministic policy. We have performed extensive experiments on several classic control and Mujoco tasks, including Hopper, Reacher, Half-Cheetah, Inverted Pendulum, Inverted Double Pendulum and Pendulum. Compared to the default DDPG in OpenAI's baseline \cite{abbeel:parameternoise}, our algorithm demonstrated substantial improvements in terms of sample efficiency. We also compared the default Gaussian exploration and the learned exploration policy and found that the exploration policy tends to visit novel states that are potentially beneficial for training the target deterministic policy.

\section{Related Work}

The idea of meta learning has been widely explored in different areas of machine learning, under different names, such as meta reinforcement learning, life-long learning, learning to learn, and continual learning. 
Some of the recent work in the setting of reinforcement learning includes \cite{duan2016rl, finn2017model, wang2016learning}, to name a few. Our work is related to the idea of learning to learn but instead of learning the optimization hyperparameters we hope to generate high quality data to better train reinforcement agents. 

Intrinsic rewards such as prediction gain \cite{bellemare:im}, learning progress \cite{Oudeyer:im}, compression progress \cite{Schmidhuber:im}, variational information maximization \cite{abbeel:vime, hester2017intrinsically}, have been employed to augment the environment's reward signal for encouraging to discover novel behavior patterns. 
One of limitations of these methods is that the intrinsic reward weighting relative to the environment reward must be chosen manually, rather than learned on the fly from interaction with the environment. Another limitation is that the reshaped reward might not guarantee the learned policy to be the same optimal one as that learned from environment rewards only \cite{ng:policyinvariance}.

The problem of exploration has been widely used in the literature. Beyond the traditional studies based on epsilon-greedy and Boltzmann exploration, there are several recent advances in the setting of deep reinforcement learning. 
For example, 
\cite{tang2017exploration} studied count-based exploration for deep reinforcement learning; 
\cite{stadie2015incentivizing} proposed a new exploration method based on assigning exploration bonuses from a concurrently learned transition model; 
\cite{hester2013learning} studied a bandit-based algorithm for learning simple exploration strategies in model-based settings; 
\cite{osband2016deep} used a bootstrapped approach for exploration in DQN, a simple algorithm in a computationally and statistically efficient manner through the use of randomized value functions \cite{osband2016random}.

\section{Reinforcement learning}
In this section, we introduce the background of reinforcement learning. 
We start with introducing Q-learning in Section~3.1, 
and then deep deterministic policy gradient (DDPG) which works for continuous action spaces in Section~3.2. 

\subsection{Q-learning}
Considering the standard reinforcement learning setting, an agent takes a sequence of actions in an environment in discrete time and collects a scalar reward per timestep. The objective of reinforcement learning is to learn a policy of the agent to optimize  the cumulative reward over future time. More precisely, we consider an agent act over  time $t\in\{1, \ldots, T\}$. At time $t$, the agent observes an environment state $s_t$ and selects an action $a_t\in A$ to take according to a policy.
 The policy can be either a deterministic function $a=\mu(s)$, or more generally a conditional probability $\pi(a|s)$. 
 The agent will then observe a new state $s_{t+1}$ and receive a scalar reward value $r_t \in R$. The set $A$ of possible actions can be discrete, continuous or mixed in different tasks. 
 Given a trajectory $\{s_t, a_t, r_t\}_{t=1}^T$, 
 the overall reward is defined as a discounted sum of incremental rewards, 
 $R=\sum_{t=1}^T \gamma^t r_t$, where $\gamma \in [0,1)$ is a discount factor. 
 The goal of RL is to find the optimal policy to maximize the expected reward. 

Q-learning~\citep{WatkinsPHD1989,WatkinsML1992} is a well-established method that has been widely used. Generally, Q-learning algorithms compute an action-value function, often also referred to as Q-function, $Q^\ast(s,a)$, which is the expected reward of taking a given action $a$ in a given state $s$, and following an optimal policy thereafter. 
The estimated future reward is computed based on the current state $s$ or a series of past states $s_t$ if available.

The core idea of Q-learning is the use of the Bellman equation as a characterization of the optimal future reward function $Q^\ast$ via a state-action-value function
\begin{equation}
Q^\ast(s_t,a) = \E [r_t + \gamma\max_{a^\prime} Q^\ast(s_{t+1},a^\prime)],
\end{equation}
where the expectation is taken w.r.t the distribution of state $s_{t+1}$ and reward $r_t$ obtained after taking action $a$. 
Given the optimal Q-function, the optimal policy greedily selects the actions with the best Q-function values. Deep Q-learning (DQN), a recent variant of Q-learning, uses deep neural networks as Q-function to automatically extract intermediate features from the state observations and shows good performance on various complex high-dimensional tasks. 

Since Q-learning is off-policy, a particular technique called ``experience replay''~\citep{LinML1992,WawrzynskiNN2009} that stores past observations from previous trajectories for training has become a standard step in deep Q-learning. 
Experience replays are stored as a dataset, also known as replay buffer, $B = \{(s_j, a_j,  r_j, s_{j+1})\}$ which contains a set of previously observed state-action-reward-future state-tuples $(s_j, a_j, r_j, s_{j+1})$. Such experience replays are often constructed by pooling such tuples generated by recent policies. 

With the replay buffer $D$, Deep Q learning follows the following iterative procedure~\citep{MnihNIPSWS2013,mnih:dqn}:  
start an episode in the initial state $s_0$; 
sample a mini-batch of tuples $M=\{(s_j, a_j, r_j, s_{j+1})\}\subseteq B$; 
compute and fix the targets $y_j = r_j + \gamma\max_a Q_{\theta^-}(s_{j+1}, a)$ 
for each tuple using a recent estimate $Q_{\theta^-}$ (the maximization is only considered if $s_j$ is not a terminal state); 
update the Q-function by optimizing the following program w.r.t the parameters $\theta$ typically via stochastic gradient descent:
\begin{equation}
\min_\theta \sum_{(s_j, a_j, r_j, s_{j+1})\in M}\left(Q_{\theta}(s_j,a_j) - y_j\right)^2.
\end{equation}
\label{eq:QFuncOpt}
Besides updating the parameters of the Q-function, 
each step of Q-learning needs to gather additional data to augment the replay buffer. 
This is done by performing an action simulation either by choosing an action at random with a small probability $\epsilon$ 
or by following the strategy $\arg\max_{a} Q_{\theta}(s_t, a)$ which is currently  estimated. 
This strategy is also called the $\epsilon$-greedy policy which is applied to encourage visiting unseen states for better exploration and avoid the training stuck at some  local minima. 
We subsequently obtain the reward $r_t$. 
Subsequently we augment the replay buffer $B$ 
with the new tuple $(s_t, a_t,r_t, s_{t+1})$ and continue until this episode terminates  or reaches an upper bound of timesteps, and then we restart a new episode. 
When optimizing w.r.t the parameter $\theta$, a recent Q-network is used to compute the target $y_j = r_j + \gamma\max_a Q_{\theta^-}(s_{j+1},a)$. 

\subsection{Deep Deterministic Policy Gradient}
For continuous action spaces, 
it is practically impossible to directly apply Q-learning, 
because the max operator in the Bellman equation, which find the optimal $a$, is usually infeasible, unless discretization is used or some special forms of Q-function are used. 
Deep deterministic policy gradient (DDPG) \citet{lillicrap:ddpg} 
addresses this issue by training a parametric policy network together with the  Q-function using policy gradient descent.  

Specifically, DDPG maintains a deterministic actor policy $\pia = \delta(a - \mu(s, \theta^\pia))$ where $\mu(s,\theta^\pia)$ is a parametric function, such as a neural network, that maps the state to actor.  
We want to iteratively update $\theta^\pia$, such that $a=\mu(s, \theta^\pia)$ gives the optimal action that maximizes the Q-function $Q(s,a)$. 
%
so that $a=  \mu(s,\theta^\pia)$ can be viewed as an approximate action-argmax operator of the Q-function, and we do not have to perform the action maximization in the  high-dimensional continuous space.
In training, the critic $Q_\theta(s, a)$ is updated using the Bellman equation as in  Q-learning that we introduced above, and the actor is updated to maximize the expected reward w.r.t. $Q_\theta(s,a)$,  
$$
\max_{\theta^\pia}  
\big \{ J(\theta^\pia) := \E_{s\sim B}[Q_\theta(s, \mu(s, \theta^\pia))] \big \}, 
$$
where $s\sim B$ denotes sampling $s$ from the replay buffer $B$. 
This is achieved in DDPG using gradient descent:  
$$
\theta^{\pia} \gets \theta^\pia + \eta \nabla_{\theta^\pia} J(\theta^\pia), 
$$
where 
\begin{align*}
\nabla_{\theta^\pia} J(\theta^\pia)  
& = \nabla_{\theta^\pia} \E_{s\sim B}[\nabla_a Q_\theta(s, \mu(s, \theta^\pia)) \nabla_{\theta^\pia} \mu(s)]. 
\end{align*}
In DDPG, the actor $\mu(s, \theta^\pia)$ and the critic $Q_\theta(s,a)$ are updated alternatively until convergence. 

As in Q-learning, 
the performance of DDPG critically depends on a proper choice of exploration policy $\pie$, which controls what data to add at each iteration. 
However, in high-dimensional continuous action space, exploration is highly nontrivial. %
In the current practice of DDPG, 
the exploration policy $\pie$ is often constructed heuristically
 by adding certain type of noise to the actor policy to encourage stochastic exploration. 
%
A common practice is to add an uncorrelated Gaussian or a correlated Ornstein-Uhlenbeck (OU) process  \cite{Uhlenbeck:ou} to the action selected by the deterministic actor policy, that is,  
$$ a =  \mu(s, \theta^\pia) ~ + ~ \mathcal{N}(0,\sigma^2).$$
Since DDPG is off-policy, the exploration can be independently addressed from the learning. It is still unclear whether these exploration strategies can always lead to desirable learning of the  deterministic actor policy.








\begin{algorithm}
\caption{Teacher: Learn to  Explore}
\label{alg:alg}
\begin{algorithmic}[1]
\STATE Initialize $\pie$ and $\pia$. 
\STATE Draw $D_1$ from $\pia$ to estimate the reward $\hat R_{\pia}$ of $\pia$. 
\STATE Initialize the Replay Buffer $B = D_1$. 
\FOR{iteration $t$} 
\STATE 
Generate $D_0$ by executing teacher's policy $\pie$.  
%
%
%
\STATE
Update actor policy $\pia$ to $\pi'$ using DDPG based on $D_0$: 
$\pi' \leftarrow \mathrm{DDPG}(\pia, D_0)$. 
\STATE
Generate $D_1$ from $\pia'$ and estimate the reward of $\pia'$. 
Calculate the meta reward: $\hat{\mathcal{R}}(D_0) = \hat R_{\pia'} - \hat R_{\pia}$. 
\STATE 
 Update Teacher's Policy $\pie$ with meta policy gradient \\
$$
\theta^{\pie} \gets \theta^{\pie} +  \eta \nabla_{\theta^{\pie}} 
\log \mathcal P(D_0 | \pie) 
\hat{\mathcal{R}}(D_0)
$$
\STATE{
Add both $D_0$ and $D_1$ into the Replay Buffer $B \gets B \bigcup  D_0 \bigcup D_1$. 
}
\STATE
Update $\pia$ using DDPG based on Replay Buffer, that is, 
$\pia \gets \mathrm{DDPG}(\pia, ~ B)$. Compute the new $\hat R_{\pia}$. 
\ENDFOR
\end{algorithmic}
\end{algorithm}

\section{Learning to Explore} 
We expect to construct better exploration strategies that are potentially better than the default Gaussian or OU exploration. In practice, e.g., in the Mujuco control tasks, the action spaces are bounded by a high-dimensional continuous cube $[-1,1]^d$. Therefore, it is very possible that the Gaussian assumption of the exploration noises is not suitable when the action selected by the actor policy is close to the corner or boundaries of this cube. Furthermore it is also possible that the actor policy gets stuck in a local basin in the state space and thus cannot escape even with random Gaussian noises added. 

All existing exploration strategies seem to be based on the implicit assumption that the exploration policy $\pie$ should stay close to the actor policy $\pia$, but with some more stochastic noise. However, this assumption may not be true. 
Instead, it may be beneficial to make $\pie$ significantly different from the actor $\pia$ in order to explore the space that has not been explored previously. 
Even in the case of using Gaussian noise for exploration, the magnitude of the Gaussian noise is also a critical parameter that may influence the performance significantly. 
Therefore, it is of great importance to develop a systematic approach to adaptively learn the exploration strategy, instead of using simple heuristics.

Since DDPG is an off-policy learning algorithm and the exploration is independent from the learning, we can decouple the exploration policy with the actor policy. We hope to construct an exploration policy which generates novel experience replays that are more beneficial for training the actor policy. 
To do so, we introduce a meta-reinforcement learning approach to learn an exploration policy so that it most efficiently improves the training of the actor policy.

%
%
%

%
 
%

\subsection{A view from MDP}
To better understand our method, we can formulate a MDP (Markov Decision Process) for the interaction between exploration agent (or \textit{teacher} with policy $\pie$) and exploitation agent (or \textit{student} with policy $\pia$). The state space $\mathcal{S}$ is defined as the collection of the (exploitation) policy $\pia$, the action in space $\mathcal{A}$ is defined as the rollouts $D_0$ generated by executing the meta-exploration-policy $\pie$. Then any \textbf{Policy Updater} could be defined as a transition function to map a policy to the next policy: 
$\mathcal{T} : \mathcal{S} \times \mathcal{A} \rightarrow \mathcal{S}$.
For example, DDPG is a off-policy \textbf{Policy Updater}. 
The reward function $\mathcal{R}: \mathcal{S} \times \mathcal{A} \rightarrow R$ 
could be defined as \textbf{Policy Evaluator} to specify the exploitation agent's performance. Furthermore, we define \textbf{meta-reward} $\mathcal{R}(D_0) = \mathcal{R}_{\pi'} - \mathcal{R}_{\pi}$ to measure the student's performance improvement. To produce a reward, for example, we can make a state transition $(\pi, D_0) \rightarrow \pi'$ with transition function \textbf{DDPG}, and get the Monte Carlo estimation of the reward $\mathcal{R}$ based on the rollouts $D_1$ generated by executing the look-ahead policy $\pi'$. For more details, please refer to Algorithm~\ref{alg:alg}.

\subsection{Learning Exploration Policy with Policy Gradient}
Our framework can be best viewed as a teacher-student learning framework, 
where the exploration policy $\pie$, viewed as the \emph{teacher}, 
generates a set of data $D_0$ at each iteration, 
and feeds it into a DDPG agent with an actor policy $\pi$ (the \emph{student}) who learns from the data and improves itself.   
Our goal is to adaptively improve the teacher $\pie$ so that it generates 
the most informative data to make the DDPG learner improve as fast as possible. 

In this meta framework, the generation of data $D_0$ can be viewed as the ``action'' taken by the teacher $\pie$, 
and its related reward should be defined as the improvement of the DDPG learner using this data $D_0$, 
%
\begin{align}\label{j0}\begin{split}
\mathcal J(\pie) 
&  = \E_{D_0\sim \pie}[\mathcal R(D_0)] \\
& = \E_{D_0\sim \pie} [R_{\mathrm{DDPG}(\pia, D_0)}  -  R_{\pia}], 
\end{split}
\end{align}
where $\pi' = \mathrm{DDPG}(\pia, D_0)$ denotes a new policy obtained from one or a few steps of DDPG updates from $\pia$ based on data $D_0$;  $R_{\mathrm{DDPG}(\pia, D_0)}$  and $R_{\pia}$ are the actual cumulative reward of rollouts generated by policies $\pi' = \mathrm{DDPG}(\pia, D_0)$ and $\pia$, respectively, in the original RL problem. 
Here we use $\mathcal R(D_0)$ to denote the ``meta'' reward of data $D_0$ in terms of how much it helps the progress of learning the agent.

Similar to the actor policy, we can parameterize this exploration policy $\pie$ by $\theta^{\pie}$. 
Using the REINFORCE trick, we can calculate the gradient of $\mathcal J(\pie)$ w.r.t. $\theta^{\pie}$: 
\begin{align}\label{metagrad}
\nabla_{\theta^{\pie}}\mathcal J = 
\E_{D_0\sim \pie} \left [ \mathcal R(D_0)  \nabla_{\theta^{\pie}} \log \mathcal P(D_0 | \pie) 
\right ] , 
\end{align}
where $\mathcal P(D_0 | \pie)$ is the probability of generating transition tuples $D_0:=\{s_t, a_t, r_t\}_{t=1}^T$ given $\pie$. This distribution can be factorized as
$$
\mathcal P(D_0 | \pie) = p(s_0) \prod_{t=0}^T \pie(a_{t}| s_t) p(s_{t+1}|s_t, a_t),  
$$
where $ p(s_{t+1}|s_t, a_t)$ is the transition probability and $p(s_0)$ the initial distribution. The dependency of the reward is omitted here. 
Because $p(s_{t+1}|s_t, a_t)$ is not involved with the exploration parameter $\theta^{\pie}$,
by taking the gradient w.r.t. $\theta^{\pie}$, we have 
$$
 \nabla_{\theta^{\pie}} \log \mathcal P(D_0 | \pie)  = 
  \sum_{t=1}^T \nabla_{\theta^{\pie}} \log \pie(a_t|s_t). 
$$
This can be estimated easily on the rollout data. We can also approximate this gradient with sub-sampling for the efficiency purpose. 

To estimate the meta-reward $\mathcal R(D_0)$, 
we perform an ``exercise move'' by running DDPG ahead for  one or a small number of steps: 
we first calculate a new actor policy $\pia' = \mathrm{DDPG}(\pia, D_0)$ by running DDPG based on data $D_0$; 
we then simulate from the new policy $\pia'$ to get data $D_1$, 
and use $D_1$ to get an estimation $\hat R_{\pia'}$ of the reward of $\pia'$. 
This allows us to estimate the meta reward by 
$$
\hat{\mathcal{R}}(D_0) = \hat R_{\pia'} - \hat R_{\pia}, 
$$
where $\hat R_{\pia}$ is the estimated reward of $\pia$, which we should have obtained from the previous iteration. 

Once we estimate the meta-reward $\mathcal R(D_0)$, we can update the exploration policy $\pie$ by following the meta policy gradient in \eqref{metagrad}. This yields the following update rule: 
\begin{align}\label{update0}
\theta^{\pie} \gets \theta^{\pie} + \eta \hat{\mathcal{R}}(D_0)  
\sum_{t=1}^T \nabla_{\theta^{\pie}} \log \pie(a_t|s_t). 
\end{align}
After updating the exploration policy, we add both $D_0$ and $D_1$ into a replay buffer $B$ that we maintain across the whole process, that is, $B\gets B\cup D_0 \cup D_1$; we then update the actor policy $\pia$ based on $B$, that is, $\pia \gets \mathrm{DDPG}(\pia, ~ B)$.  Our main algorithm is summarized in Algorithm~\ref{alg:alg}. 

It may appear that our meta update adds significant computation demand, especially in requiring to generate $D_1$ for the purpose of evaluation. However, $D_1$ is used highly efficiently since it is also added into the replay buffer and is subsequently used for updating $\pia$. This design helps improve, instead of decrease, the sample efficiency. 

Our framework allows us to explore different parametric forms of $\pie$. We tested two design choices: 

i) Similar to motivated by the traditional exploration strategy, we can set $\pie$ to equal the actor policy adding a zero-mean Gaussian noise whose variance is trained adaptively, that is,  $\pie= \normal(\mu(s, \theta^\pi), ~ \sigma^2 I)$, where $\sigma$ is viewed as the parameter of $\pie$ and is trained  with meta policy gradient \eqref{update0}. 

ii) Alternatively,  we can also take $\pie$ to be another Gaussian policy that is completely independent with $\pia$, that is, $\pie = \normal(f(s, \theta^f), \sigma^2 I)$, where $f$ is a neural network with parameter $\theta^f$, and 
$\theta^{\pie} :=[\theta^f, \sigma]$ is updated by the meta policy gradient \eqref{update0}. 

We tested both i) and ii) empirically, and found that ii) performs better than i). This may suggest that it is beneficial to explore spaces that are far away from the current action policy (see Figure~\ref{fig:contour}).  

\section{Experiments}

In this section, we conduct comprehensive experiments to understand our proposed meta-exploration-policy learning 
algorithm and to demonstrate its performance in various continuous control tasks. 
Two videos are included as supplementary material to illustrate the running results of Pendulum and Inverted Double Pendulum.


\begin{table}[t]
\centering
\begin{tabular}{l|r}
& DDPG and Meta \\\hline
Number of Epoch Cycles & 20 \\
Number of Rollout Steps & 200 \\
Number of Training Steps & 50 \\
\end{tabular}
\caption{\label{tab:trpo} Common Parameter settings for DDPG and Meta in most tasks}
\end{table}

\begin{figure}[t]
\centering
\centerline{\includegraphics[width=\columnwidth, trim={0em 2.5em 0 0},clip]{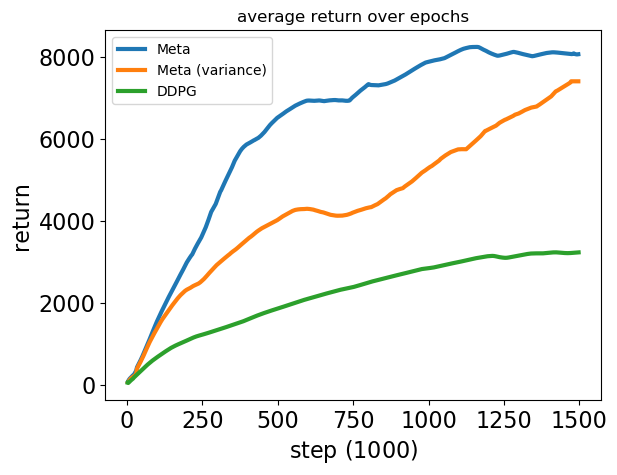}}
\rotatebox{0}{\footnotesize Time Steps ($\times 1000$)}
\caption{Comparison between meta exploration policies and DDPG}
\label{fig:behavior}
\end{figure}

\begin{figure*}[htbp]
\centering
\begin{subfigure}[b]{0.15 \textwidth}
    \includegraphics[width=\textwidth]{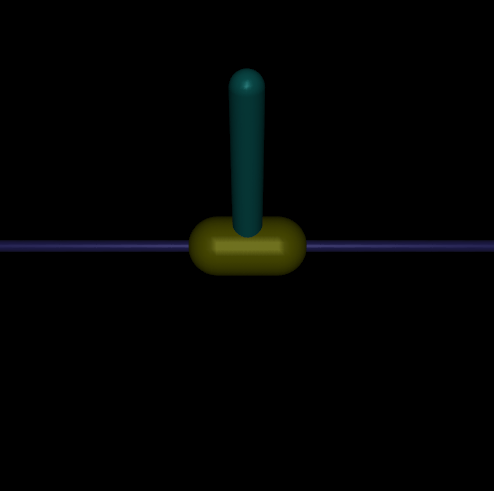}
\end{subfigure}
\begin{subfigure}[b]{0.15 \textwidth}
    \includegraphics[width=\textwidth]{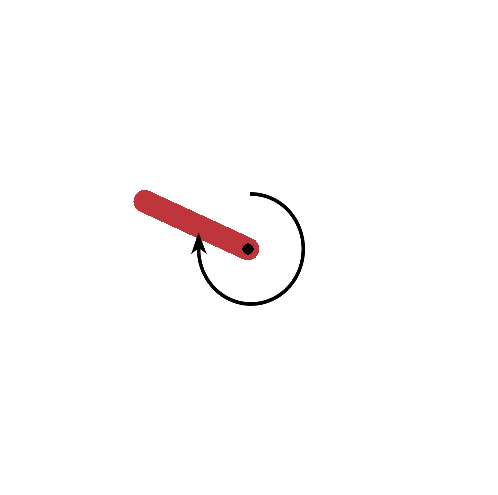}
\end{subfigure}
\begin{subfigure}[b]{0.15 \textwidth}
    \includegraphics[width=\textwidth]{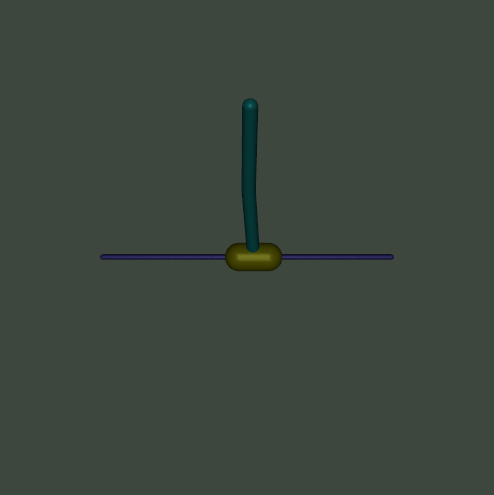}
\end{subfigure}
\begin{subfigure}[b]{0.15 \textwidth}
    \includegraphics[width=\textwidth]{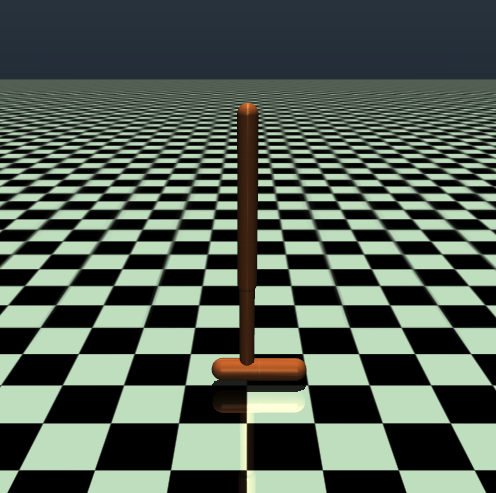}
\end{subfigure}
\begin{subfigure}[b]{0.15 \textwidth}
    \includegraphics[width=\textwidth]{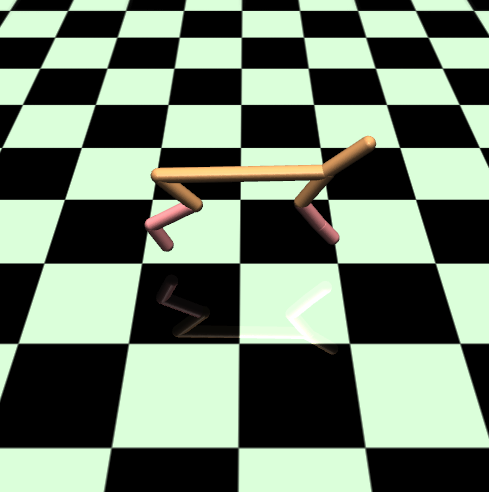}
\end{subfigure}
\begin{subfigure}[b]{0.15 \textwidth}
    \includegraphics[width=\textwidth]{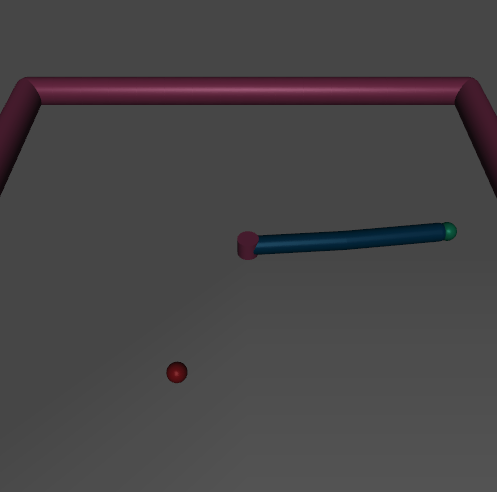}
\end{subfigure}
\caption{Illustrative screenshots of environments we experiment with Meta and DDPG}
\end{figure*}

\subsection{Experimental Setting}
Our implementation is based on the OpenAI's DDPG baseline \cite{abbeel:parameternoise} on the GitHub website\footnote{https://github.com/openai/baselines/tree/master/baselines/ddpg}. Our experiments were performed on a server with 8 Tesla-M40-24GB GPU and 40 Intel(R) Xeon(R) CPU E5-2660 v3 @ 2.60GHz processors. The deterministic actor (or student) policy network and $Q$-networks have the same architectures as implemented in the default DDPG baseline, which are multi-layer perceptrons with two hidden layers (64-64). 
For the meta-exploration policy (teacher $\pi_e$), we implemented a stochastic Gaussian policy with a mean network represented with a MLP with two hidden layers (64-64), and a log-standard-deviation variance network with a MLP with two hidden layers (64-64). 

In order to make a fair comparison with baseline, we try to set the similar hyper-parameters as DDPG. The common parameter settings for DDPG and our meta-learning algorithm in most tasks are listed in Table 1. Besides those common ones, our method has some extra parameters: exploration rollout steps (typically 100) for generating exploration trajectories, number of evaluation steps (typically 200, same as DDPG's rollout steps) for generating exploitation trajectories used to evaluate student's performance, number of training steps (typically 50, aligning with DDPG's training steps) to update student policy $\pia$, and number of exploration training steps (typically 1) to update the Meta policy $\pie$. In most experiments, we set number of cycles 20 in an epoch to align with DDPG's corresponding setting. Tasks such as Half-Cheetah, Inverted Pendulum, which need more explore rollout steps (1000) to finish the task, and ended up with 2000 evaluation steps, 500 number of training steps to update students and 100 exploration training steps to update teacher. In OpenAI's DDPG baseline \cite{abbeel:parameternoise}, the total number of steps of interactions is 1 million. Here, in tasks such as Half-Cheetah, Inverted Pendulum and Inverted Double Pendulum, it takes about 1.5 million steps, Hopper with 1 million steps, and 0.7 million and 0.9 million steps are sufficient for Reacher and Pendulum to achieve convergence. Similar to DDPG, the optimizer we use to update the network parameter is Adam \cite{kingma:adam} with the same actor learning rate $0.0001$, critic learning rate $0.001$, and additionally learning rate $0.0001$ for our meta policy. Similar to DDPG, we adopt {Layer-Normalization \cite{ba:layer-norm}} for our two policy networks and one $Q$-network.

\begin{table*}[t]
\centering
\caption{Reward achieved in different environments}
\label{table:finalreward}
\begin{tabular}{|l|l|l|}
\hline
  env-id & Meta &  DDPG  \\
  \hline  
  InvertedDoublePendulum-v1 & \textbf{7718 $\pm$ 277} & 2795 $\pm$ 1325 \\
  \hline
  InvertedPendulum-v1 & \textbf{745 $\pm$ 27} & 499 $\pm$ 23 \\
  \hline
  Hopper-v1 & \textbf{205 $\pm$ 41} & 135 $\pm$ 42 \\
  \hline
  Pendulum-v0 & \textbf{-123 $\pm$ 10} & -206 $\pm$ 31 \\
  \hline
  HalfCheetah-v1 & \textbf{2011 $\pm$ 339} & 1594 $\pm$ 298 \\
  \hline
  Reacher-v1 & -12.16 $\pm$ 1.19 & \textbf{-11.67 $\pm$ 3.39} \\
  \hline
\end{tabular}
\end{table*}

\begin{figure*}[t]
\centering
\begin{subfigure}[b]{0.3 \textwidth}
\includegraphics[width=\textwidth]{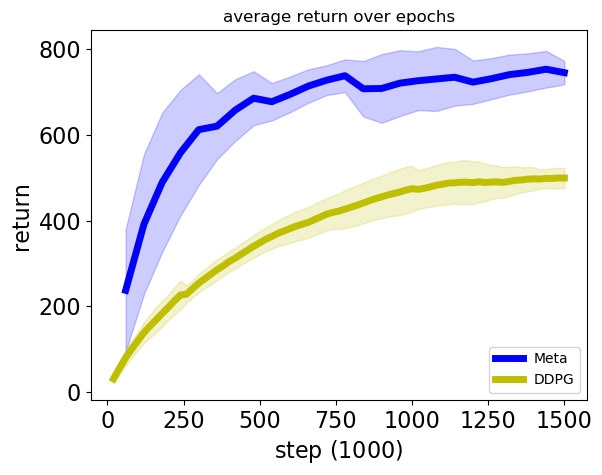}
\caption{\label{sfig:invertedpen} InvertedPendulum}
\end{subfigure}
\begin{subfigure}[b]{0.3 \textwidth}
\includegraphics[width=\textwidth]{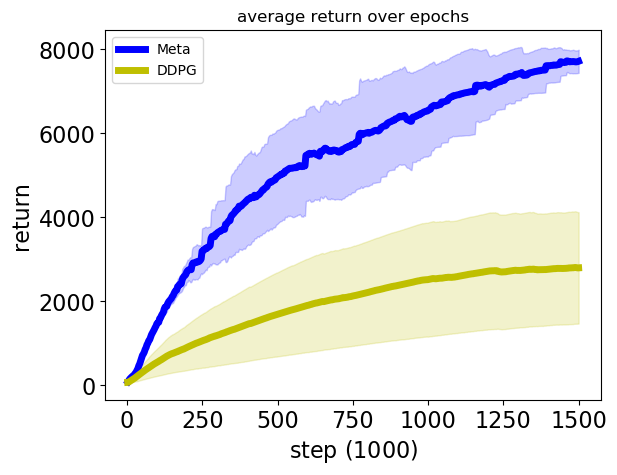}
\caption{\label{sfig:inverted} InvertedDoublePendulum}
\end{subfigure}
\begin{subfigure}[b]{0.3 \textwidth}
\includegraphics[width=\textwidth]{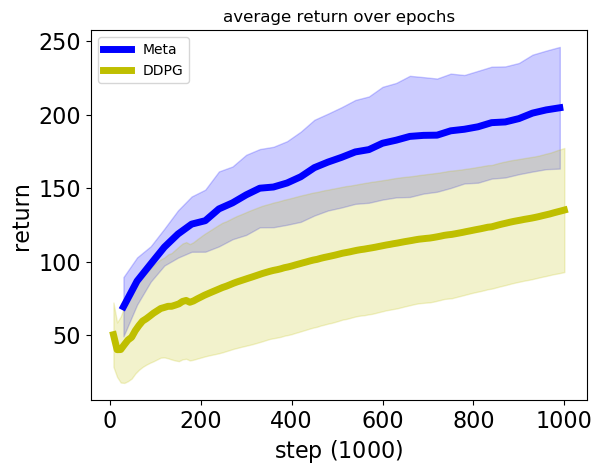}
\caption{\label{sfig:hopper} Hopper}
\end{subfigure}
\\
\begin{subfigure}[b]{0.3 \textwidth}
\includegraphics[width=\textwidth]{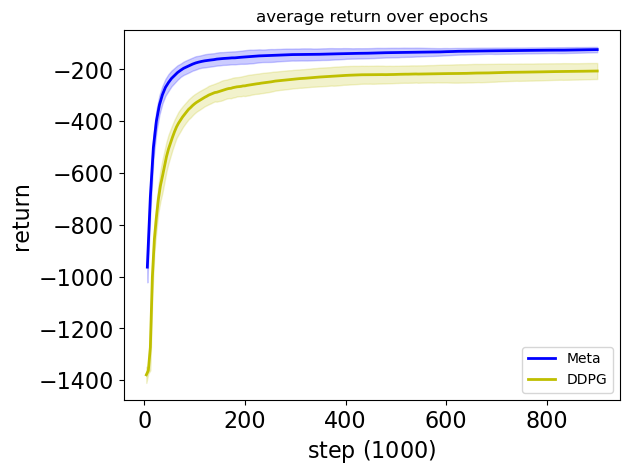}
\caption{\label{sfig:pendulum} Pendulum}
\end{subfigure}
\begin{subfigure}[b]{0.3 \textwidth}
\includegraphics[width=\textwidth]{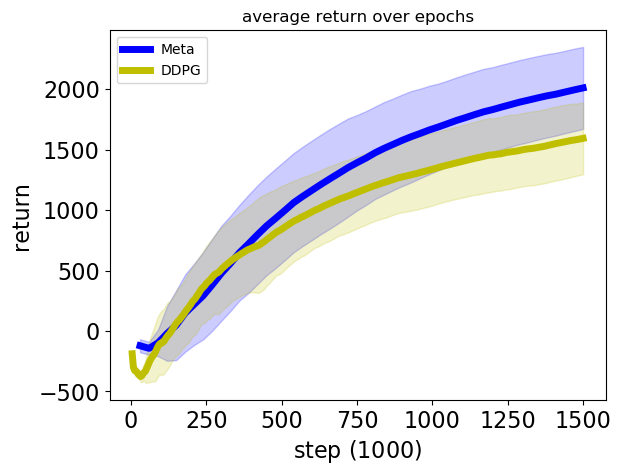}
\caption{\label{sfig:cheetah} HalfCheetah}
\end{subfigure}
\begin{subfigure}[b]{0.3 \textwidth}
\includegraphics[width=\textwidth]{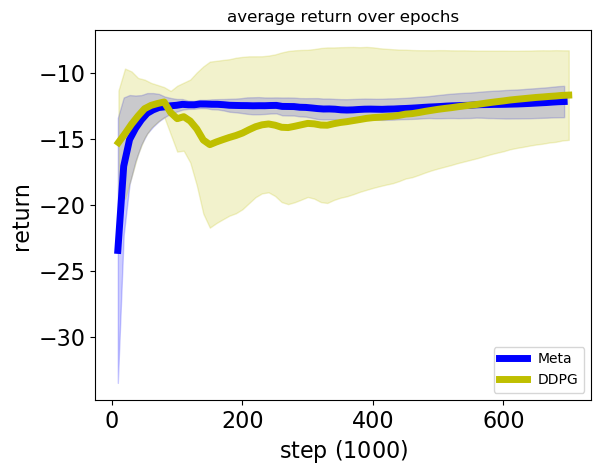}
\caption{\label{sfig:reacher} Reacher}
\end{subfigure}
\caption{Performance Comparison of Meta and DDPG for Six Continuous Control Tasks.}
\label{fig:convergence}
\end{figure*}

\subsection{ Meta Exploration Policy Explores Efficiently}
To investigate and evaluate different teacher's behaviors, we 
tested in Inverted Double Pendulum the two possible choices of policy architectures of $\pie$ listed in Section~4. 

In Figure~\ref{fig:behavior}, \texttt{Meta} denotes that 
we learn an exploration policy that is a Gaussian MLP policy with  
independent network architecture of student's policy. 
Meta runs consistently better than DDPG baseline with relative high return and  sample-efficiency. 
Usually, Meta policy learns in the same pace as student policy, it updates every time both from student's success (performance improvement) and failure (negative performance). 
For a further more robust policy updates, we may need to take consideration of the trade-off between sample efficiency and sample quality.

\begin{figure*}[t]
%
%
%
%
\begin{subfigure}[b]{0.3 \textwidth}
\includegraphics[width=\textwidth]{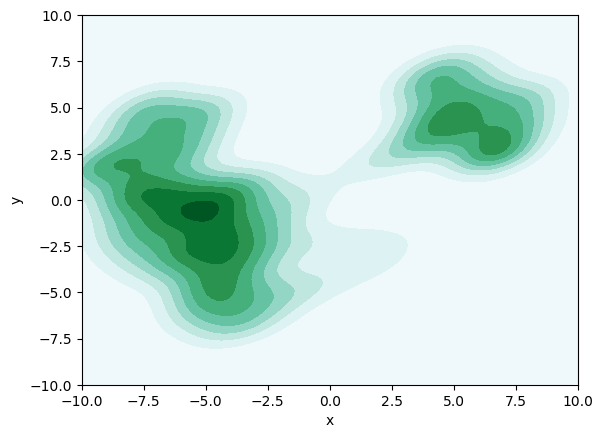}
\caption{\label{sfig:meta-teacher} Meta-Teacher (early)}
\end{subfigure}
\begin{subfigure}[b]{0.3 \textwidth}
\includegraphics[width=\textwidth]{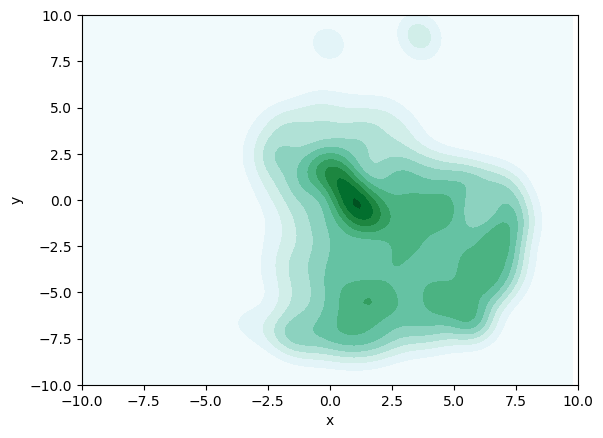}
\caption{\label{sfig:meta-student} Meta-Student (early)}
\end{subfigure}
\begin{subfigure}[b]{0.3 \textwidth}
\includegraphics[width=\textwidth]{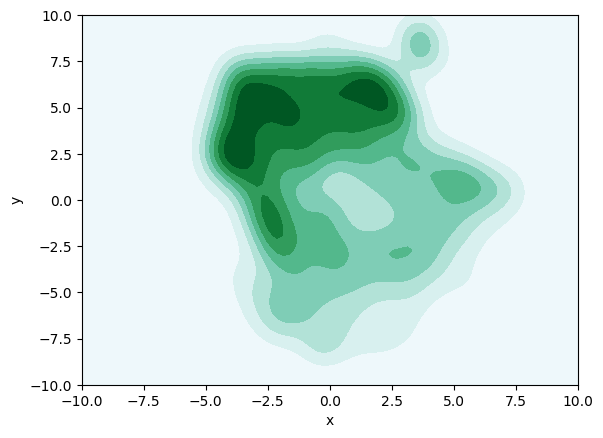}
\caption{\label{sfig:ddpg-status} DDPG (early)}
\end{subfigure}
\\
\begin{subfigure}[b]{0.3 \textwidth}
\includegraphics[width=\textwidth]{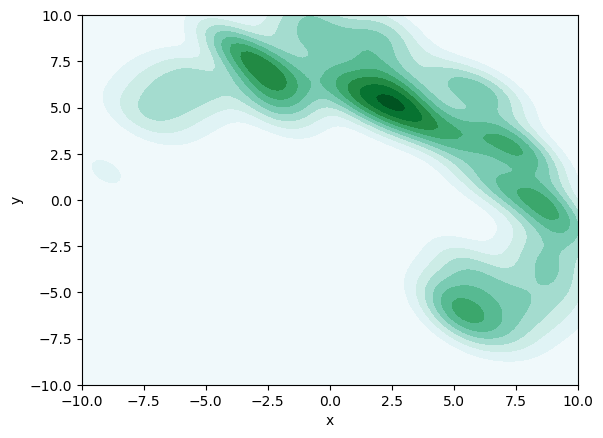}
\caption{\label{sfig:meta-teacher} Meta-Teacher (late)}
\end{subfigure}
\begin{subfigure}[b]{0.3 \textwidth}
\includegraphics[width=\textwidth]{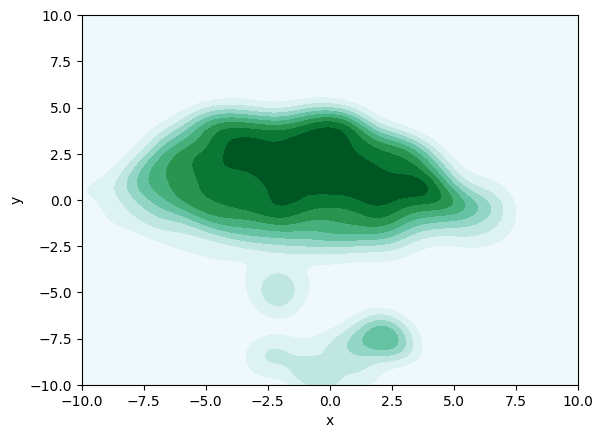}
\caption{\label{sfig:meta-student} Meta-Student (late)}
\end{subfigure}
\begin{subfigure}[b]{0.3 \textwidth}
\includegraphics[width=\textwidth]{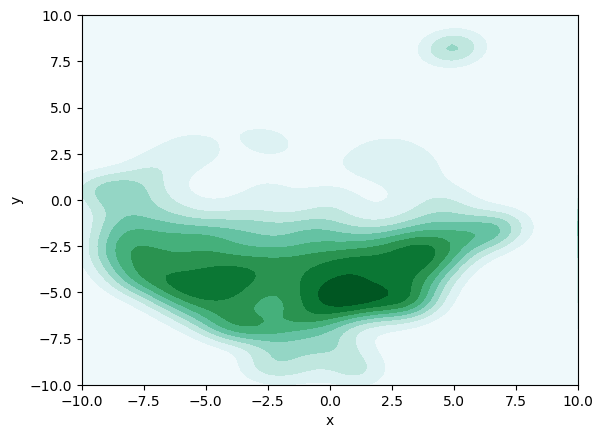}
\caption{\label{sfig:ddpg-status} DDPG (late)}
\end{subfigure}
\\
%
%

\caption{State Visitation Density Contours of Meta and DDPG in Early and Late Training Stages.} 
\label{fig:contour}
\end{figure*}

A second exploration policy denoted as \texttt{Meta (variance)} in Figure~\ref{fig:behavior} is by 
taking advantage of student's learning, combined with a variance network as $\pie = \pia + N(0, \sigma^2 I)$. Essentially, we are learning adaptive variance for exploration. Based on the student's performance, teacher is able to learn to provide training transitions with appropriate noise. This teacher's demonstrations help student to explore different regions of state space in an adaptive way.

For Figure~\ref{fig:behavior}, we can see that the fully independent exploration policy perform better than the more restrictive policy that only adds noise to the action policy. As we show in Figure~\ref{fig:contour}, the independent exploration policy tends to explore regions that are not covered by the actor policy, suggesting that it is beneficial to perform \emph{non-local exploration}. 


\subsection{Sample Efficiency in Continuous Control Tasks}
We show the learning curves in Figure~\ref{fig:convergence} for six various continuous control tasks, each is running three times with different random seeds to produce reliable comparison. Overall, our meta-learning algorithm is able to 
achieve sample-efficiency with better returns in most of the following continuous control tasks.  Significantly, in Inverted Pendulum and 
Inverted Double Pendulum, on average, in about 250 thousands out of 1500 thousands steps, we are able to achieve the similar return as the best of DDPG. That is about $1/6$ number of baseline's samples. Finally, our average return is about 7718 compared to DDPG's 2795. In 
Pendulum, we performed clearly better with higher average return, and converge faster than DDPG in less than 200 
thousand steps. In Half-Cheetah and Hopper, on average, our meta-learning algorithm is pretty robust with higher returns 
and better sample-efficiency. In Reacher, we have very similar return as 
DDPG baseline with lower variance. The possible intuition we are able to improve the sample-efficiency and returns in most of tasks is that teacher is able to learn to help student to improve their performance, which is the student's ultimate 
goal.

\subsection{Guided Exploration with Diverse Meta Policies}

To further understand the behaviors of teacher and student policies and how teacher interacts with student during the learning process, we plot the density contours of state visitation probabilities in Figure~\ref{fig:contour}. The probabilities are learned with Kernel Density Estimation based on the samples in 2D embedding space.
In Inverted Double Pendulum task, we collect about 500 thousands observation states for teacher policy and 1 million states 
for student policy. As comparison, we get 1 million states from DDPG policy. Then we project these data-sets {jointly} into 2D embedding space by {t-SNE} \cite{maaten:t-sne}. We may be able to find interesting insights, although it is possible that the {t-SNE} projection might introduce artifacts in the visualization.

As shown in Figure~\ref{fig:contour}, we have two groups of comparison studies for the evolution of teacher and student learning processes in different stages. In each row, the first column is Meta-Teacher, the second one is Meta-Student policy and the third one is the DDPG baseline. The first row (Figure~\ref{fig:contour}(a, b, c)) visualize state distributions from the first 50 roll-outs by executing the random teacher and student policies where the policies are far from becoming stationary. The bottom row (Figure~\ref{fig:contour}(d,e,f)) demonstrates the state distribution landscape visited by teacher, student and DDPG, respectively, from the last 50 roll-outs to the end of learning 

The teacher is exploring the state space in a \textit{global} way. 
In the two learning stages, the Meta-Teacher (Figure~\ref{fig:contour}(a, d)) has diversified state visitation distributions ranging from different modes in separate regions. We can see that Meta-Teacher policy has high entropy, which implies that Meta-Teacher provides more diverse samples for student. Guided by teacher's wide exploration, student policy is able to learn from a large range of state distribution regions.

Interestingly, compared to teacher's behavior, the student visits almost complementary different states in  distribution space consistently in both the early (Figure~\ref{fig:contour}(a,b)), and later  (Figure~\ref{fig:contour}(d,e)) stages. 
We can see that the teacher interacts with the student and is able to learn to explore different regions based on student's performance. Meanwhile, the student is learning from teacher's provided demonstrations and is focusing on different regions systematically. This allows the student to improve its performance consistently and continuously. It indicates that our \textit{global exploration} strategy is quite different from noise-based random walk \textit{local exploration} in principle. 

From the early (Figure~\ref{fig:contour}(b)) to the later stage (Figure~\ref{fig:contour}(e)), we find that the student is growing to be able to learn stationary and robust policies, guided by teacher's interactive exploration.
Finally, compared to DDPG (Figure~\ref{fig:contour}(f)), we achieve better return (8530 vs 2830) for this comparison, which indicates that our Meta policy is able to provide a better exploration strategy to help improve the baseline.

\section{Conclusion}

We introduce a meta-learning algorithm to adaptively learn exploration polices to collect better experience data for DDPG training. 
Using a simple meta policy gradient, we are able to efficiently improve the exploration policy and achieve significantly higher sample efficiency than the traditional DDPG training. 
Our empirical study demonstrates the significant practical advantages of our approach. 

Although most traditional exploration techniques are based on \emph{local exploration}
around the actor policy, 
we show that it is possible and more efficient to perform \emph{global exploration}, 
by training an independent exploration policy that allows us to explore spaces that are far away from the current state distribution. This finding has a substantial implication to our understanding on exploration strategies, showing that more adaptive, non-local methods should be used in order to learn more efficiently. Finally, this meta-policy algorithm is general and could be applied to the other off-policy reinforcement learning problems.

\section*{Acknowledgement}
We appreciate Kliegl Markus for his insightful discussions and helpful comments.

\bibliographystyle{icml2018}
\bibliography{metapg}

\end{document}